\documentclass[11pt,a4paper]{article}
\usepackage[hyperref]{acl2020}
\usepackage{times}
\usepackage{latexsym}

\usepackage{microtype}
\usepackage[utf8]{inputenc}
\usepackage{multirow}
\usepackage{svg}
\usepackage{float}
\usepackage{amsmath}
\usepackage{todonotes}

\usepackage{microtype}
\usepackage{booktabs, multirow} 
\usepackage{soul}

\aclfinalcopy 
\def\aclpaperid{52} 

\title{The Importance of Context in Very Low Resource Language Modeling}

\author{Lukas Edman \qquad  Antonio Toral \qquad Gertjan van Noord \vspace{.2cm}
 \\ Center for Language and Cognition \\ 
 University of Groningen \vspace{.1cm}
 \\ {\tt \small\{j.l.edman, a.toral.ruiz, g.j.m.van.noord\}@rug.nl}
}
\date{}
\begin{document}
\maketitle
\begin{abstract}
This paper investigates very low resource language model pretraining, when less than 100 thousand sentences are available. We find that, in very low resource scenarios, statistical n-gram language models outperform state-of-the-art neural models. Our experiments show that this is mainly due to the focus of the former on a local context. As such, we introduce three methods to improve a neural model's performance in the low-resource setting, finding that limiting the model's self-attention is the most effective one, improving on downstream tasks such as NLI and POS tagging by up to 5\% for the languages we test on: English, Hindi, and Turkish. 
\end{abstract}

\section{Introduction}
With the advent of the Transformer \cite{vaswani2017attention} and masked language model (MLM) pretraining \cite{devlin2018bert}, attention-based neural networks have proven quite effective at a variety of language tasks, provided that large amounts of data are available for pretraining. However, the performance can drop significantly as the number of sentences used for MLM pretraining decreases.
This poses an issue for low-resource settings such as for underrepresented languages, where there is a limited amount of monolingual data. 

Under low-resource conditions, attention-based models have difficulty learning from MLM, and as such statistical language models (SLMs) can outperform neural language models (NLMs). We demonstrate this by using a popular SLM toolkit, KenLM \cite{heafield2011kenlm}, and test its accuracy on the MLM task compared to that of a Transformer model.\footnote{The details of these tests are discussed in Section~\ref{sect:exp:init}. } The results (Table \ref{tab:acc_kenlm_xlm}) show that a trigram SLM is able to outperform the Transformer model by a wide margin for all languages when only 10 thousand sentences are available. 

\begin{table}[!htp]\centering

\scriptsize
\begin{tabular}{lrrrrr}\toprule
& &\multicolumn{3}{c}{Data Amount} \\
Language & System &10k &40k &100k \\ \midrule
\multirow{2}{*}{EN} &NLM &12.8 &30.7 & \textbf{44.6}  \\
&SLM &\textbf{29.7} &\textbf{37.9} & 42.1 \\ \midrule
\multirow{2}{*}{HI} &NLM &27.0 &\textbf{48.7} & \textbf{57.4} \\
&SLM &\textbf{45.7} &48.1 &52.4 \\\midrule
\multirow{2}{*}{TR} &NLM &6.4 &22.3 & 36.2 \\
&SLM &\textbf{23.1} &\textbf{30.5} &\textbf{39.9} \\
\bottomrule
\end{tabular}
\caption{English (EN), Hindi (HI), and Turkish (TR) MLM accuracy scores (\%) for a neural versus statistical model. }
\label{tab:acc_kenlm_xlm}
\end{table}

While an SLM might outperform a neural model on MLM, the neural model has the benefit of being easily transferable to downstream tasks by means of fine-tuning. As such, this paper seeks to determine how we can improve the performance of an NLM to that of an SLM in low-resource scenarios. We investigate three approaches:
\begin{enumerate}
    \item \textbf{Changing the input} by limiting the pretraining context size
    \item \textbf{Changing the architecture} by limiting the self-attention window
    \item \textbf{Changing the training objective} by using soft labels distilled from the SLM
\end{enumerate}

We motivate and detail these methods in Section~\ref{sect:method}, describe experiment details in Section~\ref{sect:exp}, show and discuss results in Section~\ref{sect:res}, and conclude our work in Section~\ref{sect:concl}.

\section{Methods} \label{sect:method}
When comparing the general function of an SLM to an NLM, we consider the largest difference to be the context size considered. A tri-gram SLM will consider only the context of the adjacent two words on either side. For example, the score we use for word \verb|C| in the sequence \verb|A B C D E F G| is $\log(p(C|A, B) \times p(D|B, C) \times p(E|C, D))$.

Meanwhile, self-attention allows a Transformer to consider the entire context, which in XLM is 256 tokens by default \cite{lample2019cross}. Since XLM is trained with continuous streams of text, the input size is therefore always 256, and can consist of multiple sentences.
In the low-resource setting, it may be difficult to learn important features from such a large context size. 

To tackle this, we consider three alternative approaches. First, we put the strictest limitation on context by limiting the length of the input (\S~\ref{sect:method:input}). Second, we use a limited attention scope, thereby limiting the context within the first layer of the Transformer, but allowing information to flow from larger contexts in subsequent layers (\S~\ref{sect:method:arch}). Finally, we put no explicit restriction on context size, but rather we expect the model to learn to limit itself via distillation from the limited statistical model (\S~\ref{sect:method:train}).

If context size is indeed the issue, we would expect the strictest form of limitation to perform best, as it would not need to learn to limit itself during training. This may be however too limiting for tasks which require a larger context, where we would expect that limiting attention would perform best. If context size is not the issue, we would expect that distilling knowledge from the statistical model would perform best, as its context is not limited, and the statistical model would still help the neural model learn a better strategy for language modelling than it is capable of on its own.

\subsection{Changing the Input}\label{sect:method:input}
We first limit the context size by presenting the input to a sliding window of a fixed context size. 
To stay consistent with the SLM, we only mask the middle word during MLM pretraining, padding the left and right side with \verb|BOS| and \verb|EOS| tokens respectively as needed.\footnote{We also tried just limiting the context size without changes to MLM or the input, as done in contemporary work~\citep{press2020shortformer}, but the performance was worse. }
For example, with a context size of 5 for the sentence "it is sunny today", we have:
\begin{verbatim}
    [BOS] [BOS] [MASK] is sunny
    [BOS] it [MASK] sunny today
    it is [MASK] today [EOS]
    is sunny [MASK] [EOS] [EOS]
\end{verbatim}

This approach has the benefit of a smaller input complexity and an easier training objective (since only 1 word is masked at a time). These factors should make it easier for the model to learn the importance of local context. However, as the pretraining step does not expose the model to input longer than the context size, fine-tuning with a longer context size may hurt the model's performance.

\subsection{Changing the Architecture}\label{sect:method:arch}
Rather than explicitly limiting the context, we also try limiting the model's attention towards words outside of the desired context. This is accomplished by adding a weight matrix to the query-key matrix produced during self attention. More specifically, referring to Equation 1 from \citet{vaswani2017attention}:
\begin{equation}
    \text{Attn}(Q,K,V) = \text{softmax}(\frac{QK^T}{\sqrt{d_k}})V
\end{equation}
We add a band matrix $W$ before applying softmax, where the elements within the band are 0 and the elements outside the band are $-\infty$, shown in Equation 2.\footnote{In practice we use $-10^9$.} The size of the band corresponds to the context size ($c$), as the attention scores within the band are unaffected, whereas the attention outside of the band is effectively removed. This approach is very similar to that of the Longformer \cite{beltagy2020longformer}, which has a sliding-window attention with the aim of reducing model complexity and computation in long documents.
\begin{equation*}
    \text{Attn}(Q,K,V) = \text{softmax}(W+\frac{QK^T}{\sqrt{d_k}})V,
\end{equation*}\vspace{-2ex}
\begin{equation}
    w_{i,j} = 
\begin{cases} 
  -\infty & j<i-c \\
  -\infty & j>i+c \\
  0 & \text{otherwise} 
\end{cases}
\end{equation}

While the model's self-attention range is limited to the defined context size, the ability for information from outside the context to associate with that of within the context is still possible in upper layers of the Transformer. For example, with a 6-layer encoder and a context size of 5, the first word could theoretically receive information about all words up to the 13th position. One benefit of this approach over limiting the input context (Section~\ref{sect:method:input}) is that the limitation can still be applied during fine-tuning.

\subsection{Changing the Training Objective}\label{sect:method:train}
The first two approaches have mainly been focused on the issue of context size, however the importance of an SLM having a fixed objective is not yet addressed. While an NLM still has to learn its objective, we can potentially make this easier to learn by learning from the outputs of the SLM, inspired by knowledge distillation~\cite{hinton2015distilling}. 

In the MLM task, a model is typically trained to compare its output for masked tokens to a ``hard label'', where the probability is 1 for the actual word and 0 for all others. Rather than training with hard labels, we construct a soft label from the output of the SLM. This is done by using the SLM's score of the context with each candidate word replacing the mask.\footnote{Each masked word is handled separately, so in a sentence with multiple masked words, the mask does not appear as part of the context for the SLM.} The scores are first min-max normalized, then weighted, and finally scaled to unit length (so that the probabilities sum to 1).\footnote{To limit memory usage, scores below the top 100 are zeroed out after normalization.} The weighting is done by raising each score to the $n$th power, acting as a ``hardness'' parameter, where the most likely candidates approach 1 and the least likely approach 0 as $n$ increases. We experiment with $n \in \{1, 2, 4, 6, 8\}$ and find $n=6$ to give the best results.

\section{Experimental Setup} \label{sect:exp}
We test each of our three methods on English (EN), Hindi (HI), and Turkish (TR). 
We train our models with XLM \cite{lample2019cross}, starting from a random initialization. We use the first 10 or 40 thousand sentences per language\footnote{The datasets come pre-shuffled.} from the WMT2007 NewsCrawl for English (following XLM), WMT2013 NewsCrawl for Hindi, and the WMT2016 NewsCrawl for Turkish.\footnote{\url{http://www.statmt.org/wmt16/translation-task.html}}
For all of the tests, the data is tokenized with UDPipe \cite{udpipe:2017},\footnote{We use UDPipe so that the tokenization for our POS tagging data (which comes from UD) is consistent with the pretraining.} truecased with Moses \cite{koehn2007moses}, and 10 thousand BPE \cite{sennrich2015neural} joins are used.

The architecture behind our models is a 6-layer Transformer with 8 attention heads, an embedding dimension size of 1024, dropout set at 0.1, and GELU \cite{hendrycks2016gaussian} activation. For pretraining, we use a batch size of 32, and the Adam optimizer \cite{kingma2014adam}, with a learning rate of 1e-4. We lower the learning rate to 2.5e-5 for the fine-tuning tasks. We use an early stopping criterion of no improvement in accuracy (MLM accuracy for pretraining, NLI or POS tag accuracy for fine-tuning) on the validation set for 20000 iterations, with a patience of 10. \footnote{As we used the XLM implementation from \url{https://github.com/facebookresearch/XLM}, any hyperparameters not mentioned are set at their default values.}

\subsection{Measuring MLM accuracy} \label{sect:exp:init}
For our initial experiment showing the MLM accuracy of an SLM versus an NLM, we use a trigram KenLM model as our statistical model, and XLM \cite{lample2019cross} as our neural model. Both KenLM and the XLM model are trained on the same 10 or 40 thousand sentences. Being a statistical model, KenLM's training process simply consists of tabulating frequencies, which are then used to estimate probabilities during inference. 

As KenLM outputs scores for entire sequences, we simulate prediction of a masked word by replacing the word with every word in the vocabulary, and take its highest score as its prediction.\footnote{Because these scores are chain probabilities, it is not clear how to get a perplexity score comparable to that of an NLM, which is why we chose to compare with MLM accuracy. However the MLM accuracies of the NLMs follow the same trend as their perplexities.}
We repeat this for every word in the sentence for the first 100 sentences of the dataset,\footnote{We use WMT \texttt{newstest2016} from English--German for English and English--Turkish for Turkish, and \texttt{newstest2014} for English--Hindi for Hindi.}, producing roughly 2600 examples.\footnote{Unlike in standard MLM during training, for evaluation only one token is masked in a sentence at a time. Masking multiple tokens would increase the number of queries to the KenLM model exponentially.}

\subsection{Downstream Tasks}
We fine-tune our models on the Natural Language Inference (NLI) task. For training, we use the MultiNLI dataset \cite{N18-1101}, and for development and testing, we use the XNLI dataset \cite{conneau2018xnli}. 

When fine-tuning on XNLI for our limited attention model (Section~\ref{sect:method:arch}), the first token (the \verb|CLS| token used for classification) in the final layer often cannot access information from the second sentence. As such, we instead average every token rather than simply taking the first token, which improves results dramatically. We did not find this to improve any of our results with the other approaches, so we use only the first token in the other approaches. 

We also investigate an easier task that typically requires less context, part-of-speech (POS) tagging, in appendix \ref{app:pos}. When applicable, the training data for both tasks is limited to the first 10 or 40 thousand sentences, according to the amount of data used in pretraining.

\section{Results} \label{sect:res}
We now compare the results of the SLM, normal NLM, and our 3 improvements to the NLM: limited context (NLM-C), limited attention (NLM-A), and the hybrid training objective (NLM-H). For NLM-C and NLM-A, we experiment with different context sizes and attention window sizes, ranging from 5 to 13. The SLMs are trigram models, and NLM-H uses these models for its soft labels. 
\subsection{Pretraining}
Table \ref{tab:mlm} shows the MLM accuracies for all of the methods, using 10 and 40 thousand sentences. As we can see, the standard NLM is the worst, each of the 3 additions improve on the standard NLM, with NLM-C performing similarly to the SLM.

\begin{table}[!htp]\centering
\scriptsize
\begin{tabular}{lrrrrrrrr}\toprule
& &\multicolumn{3}{c}{10k} &\multicolumn{3}{c}{40k} \\
System &Context &\multicolumn{1}{c}{EN} &\multicolumn{1}{c}{HI} &\multicolumn{1}{c}{TR} &\multicolumn{1}{c}{EN}&\multicolumn{1}{c}{HI} &\multicolumn{1}{c}{TR} \\ \midrule
NLM & 256 & 12.8 & 27.0 & 6.4 & 30.7 & 48.7 & 22.3\\\midrule
\multirow{3}{*}{NLM-C} &5 &27.4 & 45.1 & 22.4 & 37.5 & 50.1 & 31.7 \\
                       &9 &28.1 & 45.9 & 22.6 & 39.3 & \textbf{53.3}& \textbf{32.8} \\
                       &13 &29.4 & \textbf{46.2} & 22.8 & \textbf{40.4} & 52.9 & 31.0 \\ \midrule
\multirow{3}{*}{NLM-A} &5 &23.7 & 41.7 & 17.1 & 36.9 &51.5 & 30.4 \\
                       &9 &21.5 & 42.6 & 11.4 & 37.6 & 51.3& 29.7 \\
                       &13 &20.1 & 42.6 & 10.3 & 37.6 & 51.3 & 27.7 \\ \midrule
NLM-H & 256 & 22.7 & 38.9 & 14.1 & 33.1 & 48.8 & 27.6 \\ \midrule
SLM &5 &\textbf{29.7} & 45.7 & \textbf{23.1} & 37.9 & 48.1 & 30.5 \\
\bottomrule
\end{tabular}
\caption{MLM accuracies (\%), best in bold. The ``Context'' column refers to the attention window for NLM-A, and the input size for the others.}\label{tab:mlm}
\end{table}

The similarity in performance for NLM-C and SLM strongly suggests that local context is the most important factor in SLM's outperformance over NLM. This focus on local context also has an impact on the performance of rare words, as the NLM specifically fails to fill in the mask when the masked word is a word from the 80\% least frequent words. We discuss this in detail in appendix \ref{app:pret}.

NLM-A and NLM-H also outperform NLM, but not to the degree of NLM-C. While NLM-A has a similar goal as NLM-C, the degree to which information can flow from a wider context may be inhibiting the model from focusing on local context. This would explain why the accuracies decrease as the attention window increases. For NLM-H, since the context is not explicitly limited, it can similarly suffer from the complexity of self-attention.

\subsection{NLI}
Natural Language Inference (NLI), involves classifying two statements into three classes: ``contradiction'', ``entailment'', and ``neutral''. This typically would require a large context as the relation between the two sentences' meanings needs to be understood.
As our focus for two of our approaches was to limit their context, we would expect this task to be the most challenging. Our results are in Table \ref{tab:nli}.

\begin{table}[!htp]\centering
\scriptsize
\begin{tabular}{lrrrrrrrr}\toprule
& &\multicolumn{3}{c}{10k} &\multicolumn{3}{c}{40k} \\
System &Context &\multicolumn{1}{c}{EN} &\multicolumn{1}{c}{HI} &\multicolumn{1}{c}{TR} &\multicolumn{1}{c}{EN} &\multicolumn{1}{c}{HI} &\multicolumn{1}{c}{TR} \\ \midrule
NLM     & 256   & 45.6 & 41.5 & 42.0 & 53.2 & 49.8 & 49.4 \\\midrule
\multirow{3}{*}{NLM-C}                                                      &5      & 44.0 & 42.2 & 42.1 & 51.8 & 47.4 & 46.9 \\
        & 9     & 44.8 & 43.2 & 42.4 & 51.8 & 47.0 & 46.5\\
        & 13    & 45.2 & 42.5 & 41.4 & 50.1 & 47.2 & 46.5 \\ \midrule
\multirow{3}{*}{NLM-A}                                                      & 5     & 43.4 & 44.5 & 40.5 & 53.6 & 48.2 & 47.9 \\
        & 9     & 46.8 & 45.1 & 44.6 & \textbf{54.4} & \textbf{50.2} & \textbf{50.2} \\
        & 13 &\textbf{46.9} & \textbf{46.8} & \textbf{45.8} & 54.2 & 49.7 & \textbf{50.2}\\ \midrule
NLM-H   & 256   & 45.0 & 42.1 & 44.8 & 52.6 & 49.4 & 49.2 \\ 
\bottomrule
\end{tabular}
\caption{NLI accuracies (\%), best in bold. }\label{tab:nli}
\end{table}

The results on NLI differ greatly from the MLM accuracies, as NLM-A performs the best across the board, despite its MLM accuracy being lower than NLM-C (cf. Table~\ref{tab:mlm}). This is likely due to NLM-A needing no changes to the input between the pretraining and fine-tuning steps. Meanwhile, NLM-C performs more poorly as it needs to adjust to the longer input for fine tuning. 

When comparing the context sizes, we see that a larger context size in general performs better. This is in line with the idea that NLI generally demands a larger context size.

\section{Conclusion} \label{sect:concl}
Despite the ubiquity of pre-trained neural language models (NLMs) in state-of-the-art NLP, in the low-resource setting they are outperformed by statistical language models (SLMs). Their general formulation assumes a large amount of data for pretraining, so in this work we adapt them to better perform in low-resource conditions. 

We found that the complexity of self-attention on large contexts is a major inhibitor. As a solution to this, we propose shortening the attention span (NLM-A), which we show can increase the model's performance on downstream tasks. We believe that an ideal limitation of attention span would be initially very limited, but the span would increase dynamically during training. We plan to look into this further in future work.

For the best performance on MLM accuracy during pretraining itself,
we propose limiting the size of the input (NLM-C), improving upon the standard method for training neural models. This achieves SLM-level performance on the lowest resource setting (10 thousand sentences), and outperforms an SLM on slightly higher-resource settings (40 thousand sentences). In addition, the neural model with a limited context can, unlike the SLM, be transferred to downstream tasks. 

While limiting the input size (NLM-C) performs better than limiting the attention span (NLM-A) for pretraining, the opposite is the case for downstream tasks.
As a potential solution for this, we propose for future work a second pretraining step in which the non-limited input is used. 

Finally, our work primarily serves to investigate how attention-based models function with very little data. However in many real-world scenarios, transfer learning from large multilingual models is often used. Looking at the impact of these methods with multilingual transfer learning employed alongside is something we plan to do in the future.

\bibliography{custom}
\bibliographystyle{acl_natbib}

\clearpage
\appendix

\section{Pretraining Analysis} \label{app:pret}
To better understand the failures of the NLM model on MLM accuracy, we look at the performance of our models with respect to the frequency of each word during training. We split the vocabulary into 5 equal bins according to frequency and record the accuracy on those bins, shown in Table \ref{tab:freq}.
\begin{table}[!htp]\centering
\scriptsize
\begin{tabular}{lrrrrrrr}\toprule
&Bin &NLM &NLM-A &NLM-C &NLM-H &SLM \\ \midrule
\multirow{5}{*}{10k} &1 &0.0 &1.2 &14.3 &2.0 &\textbf{16.1} \\
&2 &0.0 &2.9 &9.2 &3.0 &\textbf{11.5} \\
&3 &0.1 &3.1 &9.0 &4.1 &\textbf{10.8} \\
&4 &0.0 &3.6 &9.9 &3.2 &\textbf{14.8} \\
&5 &15.4 &27.1 &32.2 &25.7 &\textbf{34.2} \\ \midrule
\multirow{5}{*}{40k} &1 &5.7 &14.8 &\textbf{17.9} &15.2 &\textbf{17.9} \\
&2 &7.5 &15.4 &10.0 &14.6 &\textbf{19.2} \\
&3 &9.9 &17.7 &19.0 &14.6 &\textbf{24.6} \\
&4 &8.5 &16.4 &17.4 &13.1 &\textbf{19.9} \\
&5 &33.5 &39.5 &\textbf{43.3} &36.2 &42.7 \\
\bottomrule
\end{tabular}
\caption{Accuracy (\%) per frequency bin for English, with bin 1 being the least frequent 20\%, and bin 5 being the most frequent 20\%. For NLM-A and NLM-C, we only report the scores for the systems with a context size of 5.}\label{tab:freq}
\end{table}

The SLM performs better across the board, but the NLM specifically fails on the least common 80\% of words when 10 thousand sentences are used. 
While less frequent, this still accounts for roughly 20\% of the words seen in training data, so the impact is understandably substantial. Interestingly, NLM-C performs similarly to SLM, which reinforces the idea that context size is the main reason why SLMs outperform standard NLMs in the low resource setting.

We also attempt to measure the ``reasonableness'' of a system's guess for MLM. 
Considering words split into multiple tokens by BPE, we measure how often the system completes them to a word that is in the vocabulary.
For example ``up@@'' could be reasonably completed with ``grade'' or ``date''. As the meaning of an entire sentence is not considered, local context is especially important for completing this task. We show the results in Table \ref{tab:word_comp}.

\begin{table}[!htp]\centering
\scriptsize
\begin{tabular}{lrrrrrrr}\toprule
& &NLM &NLM-A &NLM-C &NLM-H &SLM \\ \midrule
\multirow{2}{*}{10k} &EN &2.2 &22.8 &52.8 &33.9 &\textbf{61.1} \\
&TR &4.4 &32.2 &\textbf{42.2} &32.8 &39.9 \\ \midrule
\multirow{2}{*}{40k} &EN &40.3 &57.3 &69.7 &57.0 &\textbf{78.7} \\
&TR &45.3 &54.7 &55.0 &51.9 &\textbf{55.1} \\
\bottomrule
\end{tabular}
\caption{Word completion (\%) for English and Turkish. Showing systems with context 5 for NLM-A and NLM-C.}\label{tab:word_comp}
\end{table}

The results show a drastic difference in performance of NLM to SLM when trained on 10 thousand sentences. The standard NLM seems to fail to understand the concept of multi-token words. NLM-C and SLM again perform similarly. Interestingly, the discrepancy in performance on the two languages for the SLM is larger than for the NLMs. While this not central to the topic of this paper, it may be worth exploring it further.

Despite performing well on the downstream tasks, NLM-A does not perform particularly well on these pretraining metrics. This may showcase the inherent difficulty in evaluating the quality of the pretraining objective, as metrics like MLM accuracy or word completion do not give a clear indication of the transferability of a pretrained model to a downstream task. 

\section{POS Tagging} \label{app:pos}
Part-of-speech (POS) tagging is considered a much easier task than NLI, as most words do not need a large amount of context to be tagged. This should be an ideal setting for the context-limited methods to perform well, particularly NLM-C.

We use the POS tagging data from Universal Dependencies (UD) v2.7 \cite{11234/1-3424}, using the English-GUM and Turkish-BOUN datasets.

The results on POS tagging (Table \ref{tab:pos}) are somewhat similar to the NLI results, as NLM-A again performs the best. As this task is more suited for the contextually-limited NLM-C, we would expect it to perform similarly well, however this is not the case. We believe NLM-C's poor performance can again be attributed to the increase in context size for fine-tuning. 

\begin{table}[!htp]\centering
\scriptsize
\begin{tabular}{lrrrrrr}\toprule
& &\multicolumn{2}{c}{10k} &\multicolumn{2}{c}{40k} \\
System &Context &\multicolumn{1}{c}{EN} &\multicolumn{1}{c}{TR} &\multicolumn{1}{c}{EN} &\multicolumn{1}{c}{TR} \\ \midrule
NLM & 256 &89.2 & 87.5 & 92.8 & 88.9 \\\midrule

\multirow{3}{*}{NLM-C} &5 &90.8 & 87.2 & 91.7 & 87.7 \\
&9 &90.5 & 87.7 & 92.1 & 88.4 \\
&13 &90.7 & 88.3 & 92.2 & 88.2 \\ \midrule
\multirow{3}{*}{NLM-A} &5 &\textbf{92.5} & \textbf{89.2} & 94.2 & 90.0 \\
&9 &\textbf{92.5} & \textbf{89.2} & 93.9 & \textbf{90.1}\\
&13 &91.6 & 88.5 & \textbf{94.3} & 90.0\\ \midrule
NLM-H & 256 & 91.4 & 88.3 & 93.1 & 88.9 \\ 
\bottomrule
\end{tabular}
\caption{POS tagging accuracies (\%), best in bold.}\label{tab:pos}
\end{table}

The importance of local context for the POS tagging task is highlighted by the scores of NLM-A and NLM-C, where overall the models with a smaller context perform better than those with a larger context. NLM-H however does still provide improvements over the standard NLM, which may indicate that the network can more easily learn to limit its self-attention from the soft labels.

\end{document}